\documentclass[conference]{IEEEtran}
\IEEEoverridecommandlockouts
\usepackage{cite}
\usepackage{amsmath,amssymb,amsfonts}
\usepackage{algorithmic}
\usepackage{graphicx}
\graphicspath{{figures/}}
\usepackage{hyperref}
\usepackage{textcomp}
\usepackage{multirow}
\usepackage{booktabs}
\usepackage{xcolor}
\def\BibTeX{{\rm B\kern-.05em{\sc i\kern-.025em b}\kern-.08em
    T\kern-.1667em\lower.7ex\hbox{E}\kern-.125emX}}
\begin{document}
\nocite{b1,b2,Chataut2024,b3,b4,b5,b6,b7,b8,b9,b10,b11,ghimire2025cnnsoutperformtransformersmambas,jha2026dentiaskvqabenchmarkmultimodal,adhikari2026securing}

\title{A Comparative Study of GAN-Based Deep Learning Models for Pneumonia Detection in Chest X-Rays}

\author{
\IEEEauthorblockN{
Roshan Paudel\IEEEauthorrefmark{1},
Aashish Ghimire\IEEEauthorrefmark{1},
Pramod Acharya\IEEEauthorrefmark{1}
}
\IEEEauthorblockA{
\IEEEauthorrefmark{1}Department of Computer Science, University of South Dakota\\
\texttt{roshan.paudel@coyotes.usd.edu, aashish.ghimire@coyotes.usd.edu,} \\
\texttt{pramod.acharya@coyotes.usd.edu}
}
}

\maketitle

\begin{abstract}
This study evaluates pneumonia classification in chest X-rays using VGG19, MobileNetV2, ResNet50, and a custom CNN, and explores Generative Adversarial Network (GAN)-based synthetic data augmentation. MobileNetV2 achieved the highest reported accuracy of 88\% with balanced class-wise performance. The custom CNN achieved pneumonia recall of 92.67\% and precision of 79.43\%, highlighting a precision--recall trade-off. Accuracy, F1-score, precision, recall, confusion matrices, and training curves were used to assess performance.

Synthetic pneumonia images were combined with real images to investigate whether augmentation could improve classification performance. In the reported VGG19 comparison, augmented-data training accuracy reached approximately 100\%, while validation accuracy remained near 50\%, below the real-data validation accuracy. This experiment therefore did not demonstrate a validation-performance benefit from GAN augmentation. The classifier comparison highlights differences in accuracy and pneumonia recall, while the augmentation experiment indicates the need for further evaluation of synthetic-image quality and training settings.

\end{abstract}

\begin{IEEEkeywords}
GAN, Data Augmentation, Pneumonia Detection, Deep Learning Models
\end{IEEEkeywords}

\section{\textbf{Introduction}}

Pneumonia is a serious health problem worldwide, and its timely diagnosis is very important for effective treatment. Chest X-rays are the first line of diagnosis, but the performance of any machine-learning model is limited by the availability of labeled medical data, especially positive pneumonia cases. Traditional augmentation techniques, such as rotation and flipping of images, are usually used to artificially increase the size of a dataset. However, these methods may not capture all clinically relevant variation, motivating investigation of additional augmentation strategies \cite{b1}.

Recent advances have explored the use of GANs to generate synthetic medical images, but achieving high-resolution outputs with diagnostic accuracy remains a challenge. This project investigates GAN-based generation of synthetic pneumonia chest X-rays as a possible approach to supplementing real training data. The intended benefit is to introduce additional image variation; whether this improves generalization must be assessed experimentally. SSIM and PSNR can assess aspects of image similarity, but do not by themselves establish diagnostic validity or usefulness as training data \cite{b1},\cite{b2}.
Recent studies have also investigated alternative representations of chest X-ray images for pulmonary abnormality screening. For example, Chataut et al. \cite{Chataut2024} explored shape-aware thoracic edge-map representations for pulmonary abnormality classification, demonstrating that complementary structural and texture information can be useful for chest X-ray analysis.

This study compares a custom CNN with transfer learning-based MobileNetV2, VGG19, and ResNet50 models and explores the effect of synthetic data augmentation. These architectures provide different approaches to image classification. Integrating synthetic data with real images is intended to improve training-data variation, with performance assessed using accuracy, precision, and recall. The reported experiments examine both the classification performance of the selected models and the limitations of the implemented GAN augmentation approach \cite{b3}.


\begin{figure*}[ht]
    \centering
    \includegraphics[width=\textwidth]{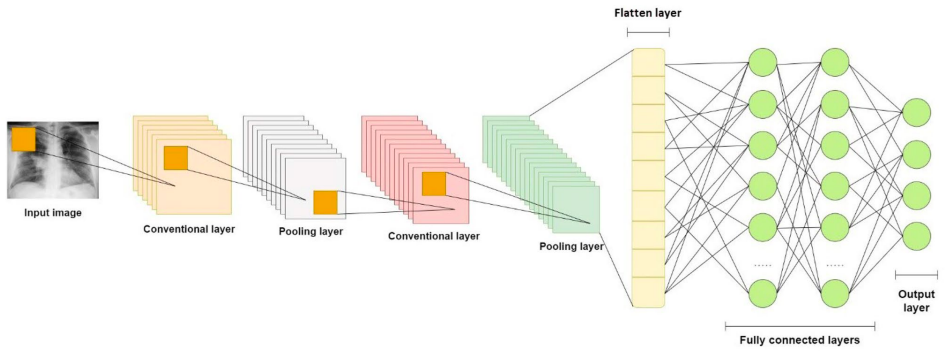}
    \caption{Shown is a generic architecture of a sequential CNN applied for chest X-ray pneumonia detection.}
    \label{fig:cnn_architecture}
\end{figure*}

\begin{figure}[htbp]
    \centering
    \includegraphics[width=0.3\textwidth]{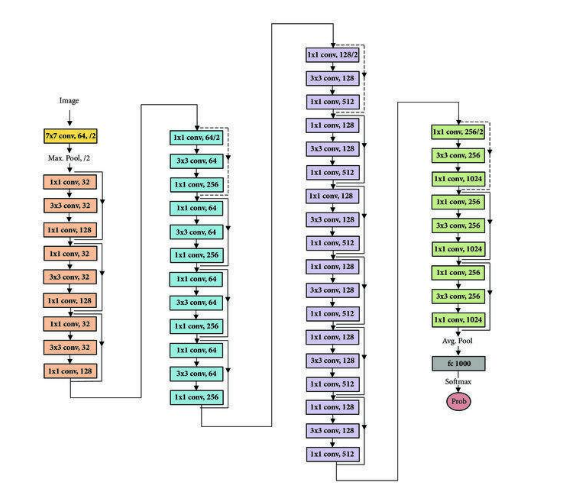}
    \caption{Shown is a Resnet50 Architecture}
    \label{fig:arch-resnet}
\end{figure}

\begin{figure}[htbp]
    \centering
    \includegraphics[width=0.3\textwidth]{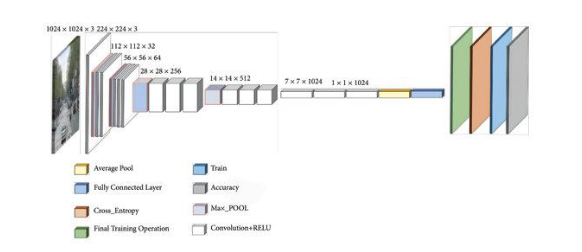}
    \caption{Shown is a MobileNet Architecture}
    \label{fig:arch-mobilenet}
\end{figure}

\begin{figure}[htbp]
    \centering
    \includegraphics[width=0.3\textwidth]{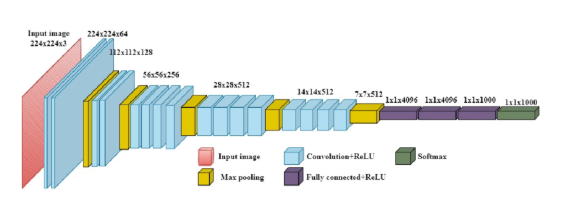}
    \caption{Shown is a VGG19 Architecture}
    \label{fig:arch-vgg}
\end{figure}

\section{\textbf{Related Work}}

Several works have sought to solve the problem of few medical images by applying a data augmentation technique. Traditional approaches for image augmentation, which include rotation, flipping, and cropping, have been implemented. These methods provide additional image variation, although their ability to represent subtle disease features may be limited. Their effectiveness therefore depends on the dataset and classification task.

GANs have received considerable attention in recent times in terms of medical image augmentation. A recent approach was proposed by Albahli et al., which used GAN to augment chest radiographs for diagnosing COVID-19 pneumonia. This model demonstrated promising results in synthesizing images that were similar to real X-rays. Similarly, other works of Sundaram and Hulkund have focused on GAN-based data augmentation for the classification of chest X-ray images. It showcased the increased classification performance enabled by a more diverse set of training images. In such studies, synthetic-image generation involves a challenging training process, and measures such as SSIM and PSNR assess image similarity rather than establish equivalent diagnostic quality \cite{b3},\cite{b4},\cite{b5}.

Comparative evidence from other radiographic tasks also motivates evaluating established CNN architectures. Ghimire et al. \cite{ghimire2025cnnsoutperformtransformersmambas} compared CNN, transformer, and Mamba models for dental caries segmentation and found that DoubleU-Net achieved the highest Dice score under their evaluation protocol. Although the task differs from pneumonia classification, this finding supports assessing architectures empirically for the target dataset rather than assuming that newer model families will perform better.

In this work, custom CNN and transfer learning models (MobileNetV2, ResNet50, and VGG19) were used for the classification of pneumonia on chest X-rays. Transfer learning models leveraged pre-trained weights from the ImageNet dataset to support feature extraction. The weights of the backbone layers in these models were frozen, while the output layers were trained for the pneumonia detection task. Output layers were designed to have only one unit, using the \textbf{sigmoid activation function} for image classification between two classes ("Normal" and "Pneumonia") and making the output consistent with binary cross-entropy loss.

Hyperparameter tuning is an important aspect of any machine learning task in order to maximize performance. Among important hyperparameters, \textbf{optimizer} refers to a key one; accordingly, in this work, Adam is employed—a method with adaptive learning rates—along with all models trained on a learning rate of 0.001. The \textbf{ReLU activation function} was used in convolutional and dense layers for effective nonlinear feature extraction, while \textbf{sigmoid} was used only in the output layer for binary classification. \textbf{Dropout} of 0.5 was used in the custom CNN to limit overfitting. \textbf{Pooling layers}---MaxPooling2D for the custom CNN and GlobalAveragePooling2D for transfer learning models---were used to reduce dimensionality. All models were trained for \textbf{15 epochs} using the \textbf{binary cross-entropy loss function}, a standard choice for binary classification.

This configuration used pretrained features while adapting the classification output to pneumonia detection. The reported results show differences in accuracy, precision, and recall across models, rather than uniformly balanced performance.

\section{\textbf{Data}}

The primary dataset for this project is sourced from Kaggle’s Chest X-ray Dataset, which contains 5,863 labeled images categorized as “Normal” or “Pneumonia.” These images specifically focus on pediatric patients aged 1-5 and were collected from the Guangzhou Women and Children’s Medical Center. The data quality is assured through review by two expert physicians, with a third ensuring reliable diagnostic labeling. This robust quality assurance process ensures that the data is suitable for machine learning model training \cite{b6}.

The dataset is split into training, testing, and validation sets for model development and evaluation. Synthetic images were generated using a Generative Adversarial Network (GAN) to supplement the available pneumonia images. SSIM and PSNR are discussed as image-similarity measures, but numerical results are not reported in this paper, so synthetic-image quality cannot be established from these measures here. Combining real and synthetic images was intended to increase training-data variation; its effect on classification performance is examined in the reported experiments. The classification reports contain 300 evaluated images, with 150 images per class.

\begin{figure}[h!]
    \centering
    \includegraphics[width=0.5\textwidth]{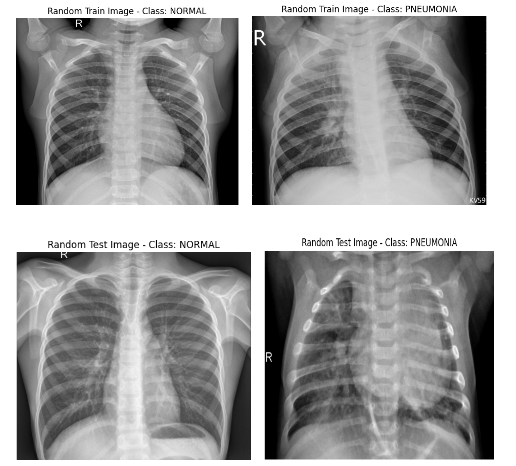}
    \caption{Images showing x-ray image (pneumonia and normal image)}
    \label{fig:data-only}
\end{figure}

\section{\textbf{Methodology}}

The methodology of this project consists of a multistage pipeline of creating synthetic pneumonia-positive chest X-ray images, integrating these with real images, and training deep learning models for the detection of pneumonia. Data preprocessing, therefore, includes cleaning and resizing the images; GAN-based generation of synthetic pneumonia images; integration of the dataset, merging real and synthetic images into one augmented dataset. The pipeline was designed to investigate whether synthetic images could supplement the available training data. A wide variety of deep learning models have been used in classifying chest X-rays into "Normal" and "Pneumonia" classes: custom CNNs and transfer learning-based architectures such as MobileNetV2, ResNet50, and VGG19. Transfer learning models were initialized with pre-trained weights, and the classifiers used a sigmoid output for binary classification. Performance was measured with accuracy, precision, recall, F1-score, and confusion matrices \cite{b7}.

\subsection{\textbf{Data Preprocessing}}

The Chest X-ray dataset included both positive pneumonia and normal cases, preprocessed to uniform dimensions and quality by resizing the images to 512 $\times$ 512~pixels. This standardizes the input dimensions for GAN training and classification. Normalization of pixel values was done to have better convergence of the model and improve the performance of GAN in generating realistic images. The real dataset was divided into training, validation, and test sets.

\subsection{\textbf{Classification Models}}

For our project, we utilized the following models: CNN, MobileNet, VGG19, and ResNet50. CNN provides basic functionalities for image processing, while MobileNet offers efficiency if used on mobile devices. VGG19 and ResNet50 feature deeper architectures, which contribute to high performance in image recognition tasks \cite{b7},\cite{b8},\cite{b9}. Semantic segmentation approaches have also demonstrated the effectiveness of convolutional architectures for pixel-level visual analysis. Ghimire et al. \cite{Ghimire2022} evaluated multiple semantic segmentation architectures for automated defect localization and reported the effectiveness of lightweight encoder-decoder approaches using IoU and F1-score metrics. CNN robustness to input perturbations has also been studied outside the medical-imaging domain: Wang et al. combined attention modules with non-local blocks to improve resilience to adversarial and noisy inputs \cite{10835588}, and later proposed symmetry-invariant CNN operations to further enhance robustness and generalization \cite{Wang2026}.

\begin{itemize}
    \item \textbf{CNN:} A Convolutional Neural Network (CNN) is a kind of deep learning model developed specifically for the analysis of visual data. It automatically extracts the spatial hierarchies and features from images, such as edges and textures, through convolutional layers, allowing it to perform tasks like image classification and object detection effectively. CNNs include multiple layers, such as convolutional, pooling, and fully connected layers, that together contribute to reducing the dimensionality, capturing the important features, and making accurate predictions. This architecture makes CNNs especially efficient in handling large-scale visual data \cite{b3},\cite{b7}. \\
    
    \item \textbf{MobileNet:} MobileNet is a family of lightweight CNNs introduced by Google, intended for mobile and embedded vision applications. It employs depthwise separable convolutions that reduce the number of parameters and computational load significantly, hence it's ideal for resource-constrained real-time tasks like image classification, object detection, and semantic segmentation on mobile devices. MobileNets have a small model size, low computational cost, and high performance, thereby guaranteeing competitive accuracy despite being lightweight. They support a host of real-world applications, which range from mobile and embedded devices to real-time analytics and edge computing. Variants such as MobileNetV2 and MobileNetV3 optimize the performance even more by including enhancements like inverted residual blocks and neural architecture search techniques \cite{b5},\cite{b9}.\\
    
    \item \textbf{VGG19:} VGG19 is a 19-layer deep convolutional neural network proposed by the Visual Geometry Group of the University of Oxford. This network is famous for its simplicity and depth, relying on the use of small 3$\times$3 convolution filters that increase in numbers progressively in order to model complex image patterns. Though computationally expensive due to the number of parameters it uses, VGG19 yields excellent performance on image classification tasks. This is a 16-layer convolutional architecture with 3 fully connected layers at the back, in which it uses a simple and repeated structure, making the model much easier to understand and implement. The 3$\times$3 convolutional layers, ReLU activation functions, max pooling with 2$\times$2 filters, and softmax layer in the output constitute this \cite{b10}.\\
    
    \item \textbf{ResNet50:} ResNet50 is a convolutional neural network belonging to the ResNet family created by Microsoft, which contains 50 layers. It introduced "residual connections" or skip connections, enabling gradients to flow through the network, thereby addressing the vanishing gradient problem and thus allowing very deep networks to be trained. ResNet50 is effective in recognizing images and, at the same time, computationally more efficient compared to other deep networks that are without residual connections. It consists of residual blocks with two or three convolutional layers with batch normalization and ReLU activation, respectively, and a shortcut connection that skips one or two layers. This architecture allows ResNet50 to train much deeper networks without performance degradation, allowing for improved accuracy and efficiency for a wide range of computer vision tasks \cite{b2},\cite{b8},\cite{b10}.\\
\end{itemize}

\subsection{Synthetic Data Generation Using GANs}
\begin{figure}[htbp]
    \centering
    \includegraphics[width=0.3\textwidth]{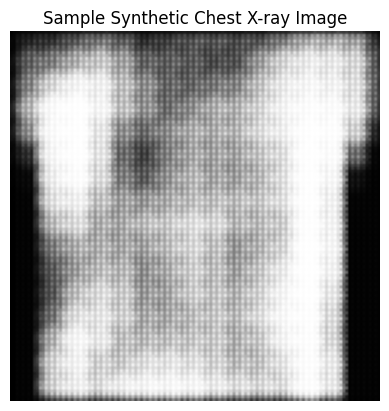}
    \caption{Shown is a sample of a synthetic chest X-ray image.}
    \label{fig:synthetic}
\end{figure}

The GAN was trained on a subset of chest X-rays labeled as pneumonia to generate synthetic images. Its architecture consists of a generator network and a discriminator network. The generator was trained to produce images resembling the training examples, while the discriminator learned to distinguish real from synthetic images. This adversarial process aims to improve image realism, but the sample shown contains visible artifacts and blurred detail. The reported outputs do not establish that the generated images preserve diagnostically relevant pneumonia features.

These synthetic pneumonia-positive images were then combined with the original real dataset to create an augmented dataset comprising both real and synthetic data. In this regard, the goal was to help alleviate data sparsity and investigate whether additional training images could improve classifier performance.

The GAN architecture consists of two main components:
\begin{itemize}
    \item \textbf{Generator:} The generator network learns to produce synthetic X-ray images that resemble real pneumonia-positive images. The network is trained to approximate characteristics of the training data, including intensity, texture, and structure.
    \item \textbf{Discriminator:} The discriminator network learns to distinguish between real and synthetic images. It is trained to classify images as either real (from the dataset) or fake (generated by the generator). The goal of the GAN is to improve the generator’s ability to produce realistic images while making it difficult for the discriminator to differentiate between real and fake images.
\end{itemize}

The generator and discriminator networks are trained together in an adversarial process, with the generator receiving feedback from the discriminator. Structural Similarity Index (SSIM) and Peak Signal-to-Noise Ratio (PSNR) assess structural and pixel-wise similarity, respectively. Numerical SSIM and PSNR results and the reference-image pairing procedure are not provided here; consequently, these measures cannot support a claim of validated synthetic-image quality in the present report.

\subsection{\textbf{Model Training and Transfer Learning}}

The classification experiments used a custom CNN and transfer learning models, including MobileNetV2, VGG19, and ResNet50. The reported VGG19 training curves compare real-data and GAN-augmented training. The transfer learning models used pretrained backbones for feature extraction and a trainable output layer for binary classification. The hyperparameter tables summarize the reported settings; a separate hyperparameter-search procedure is not documented.

As specified in the hyperparameter tables, the pretrained backbone layers were frozen and only the final dense layer was trained. This configuration adapts the classification head rather than fine-tuning the backbone. ImageNet weights supplied the initial feature representation for the transfer learning models \cite{b9},\cite{b10}.

\begin{table}[h!]
\centering
\caption{Tuned Hyperparameters for Custom CNN and MobileNetV2 Models}
\begin{tabular}{|l|l|}
\hline
\textbf{Hyperparameter}   & \textbf{Custom CNN}                                           \\ \hline
Base Architecture         & None                                                        \\ \hline
Optimizer                 & Adam                                                        \\ \hline
Learning Rate             & 0.001                                                      \\ \hline
Activation Functions      & ReLU (Conv, Dense), Sigmoid (Output)                        \\ \hline
Dropout                   & 0.5                                                        \\ \hline
Pooling                   & MaxPooling2D                                               \\ \hline
Epochs                    & 15                                                         \\ \hline
Loss Function             & Binary Crossentropy                                        \\ \hline
Trainable Layers          & All                                                        \\ \hline
\end{tabular}

\vspace{0.5cm} 

\begin{tabular}{|l|l|}
\hline
\textbf{Hyperparameter}   & \textbf{MobileNetV2}                                        \\ \hline
Base Architecture         & MobileNetV2                                                \\ \hline
Optimizer                 & Adam                                                        \\ \hline
Learning Rate             & Default (0.001)                                            \\ \hline
Activation Functions      & ReLU (Backbone), Sigmoid (Output)                          \\ \hline
Dropout                   & None                                                       \\ \hline
Pooling                   & GlobalAveragePooling2D                                     \\ \hline
Epochs                    & 15                                                         \\ \hline
Loss Function             & Binary Crossentropy                                        \\ \hline
Trainable Layers          & Final Dense Layer                                          \\ \hline
\end{tabular}
\label{tab:custom_mobilenetv2}
\end{table}

\begin{table}[h!]
\centering
\caption{Tuned Hyperparameters for ResNet50 and VGG19 Models}
\begin{tabular}{|l|l|}
\hline
\textbf{Hyperparameter}   & \textbf{ResNet50}                                           \\ \hline
Base Architecture         & ResNet50                                                  \\ \hline
Optimizer                 & Adam                                                      \\ \hline
Learning Rate             & Default (0.001)                                          \\ \hline
Activation Functions      & ReLU (Backbone), Sigmoid (Output)                        \\ \hline
Dropout                   & None                                                     \\ \hline
Pooling                   & GlobalAveragePooling2D                                   \\ \hline
Epochs                    & 15                                                       \\ \hline
Loss Function             & Binary Crossentropy                                      \\ \hline
Trainable Layers          & Final Dense Layer                                        \\ \hline
\end{tabular}

\vspace{0.5cm} 

\begin{tabular}{|l|l|}
\hline
\textbf{Hyperparameter}   & \textbf{VGG19}                                             \\ \hline
Base Architecture         & VGG19                                                    \\ \hline
Optimizer                 & Adam                                                    \\ \hline
Learning Rate             & Default (0.001)                                        \\ \hline
Activation Functions      & ReLU (Backbone), Sigmoid (Output)                      \\ \hline
Dropout                   & None                                                   \\ \hline
Pooling                   & GlobalAveragePooling2D                                 \\ \hline
Epochs                    & 15                                                     \\ \hline
Loss Function             & Binary Crossentropy                                    \\ \hline
Trainable Layers          & Final Dense Layer                                      \\ \hline
\end{tabular}
\label{tab:resnet_vgg19}
\end{table}

\subsection{\textbf{Advantages of Selected Hyperparameters and Potential Improvements}}

The hyperparameters for the models have been chosen in order to balance performance, computational efficiency, and the prevention of overfitting. \textbf{Optimizers}: Adam incorporates the advantages of both momentum and adaptive learning rates and works very well for deep learning models, especially in the case of noisy gradients or sparse data. \textbf{Learning rate}: 0.001 was the reported setting; the results do not establish that this value was optimal. Activation functions include ReLU for the hidden layers and Sigmoid for the output layers; these allow non-linear modeling while ensuring that outputs are bounded for binary classification tasks. In the Custom CNN, a dropout rate of 0.5 helps prevent overfitting by randomly shutting down neurons during training to enforce generalization. These are followed by pooling strategies: \textbf{MaxPooling2D} for CNN and \textbf{GlobalAveragePooling2D} for the pre-trained architectures, which include MobileNetV2, ResNet50, and VGG19—these reduce the dimensionality of feature maps while retaining significant features. \textbf{Binary crossentropy} is naturally fitted to binary classification problems and provides well-defined gradients for optimization. Finally, tuning of trainable layers, for instance, training all layers in the Custom CNN compared to training only the last dense layers in pre-trained models, balances computational cost and performance in leveraging transfer learning \cite{b11}.

\subsection{\textbf{Potential Hyperparameter Tuning}}

Potential improvements could be investigated by varying other hyperparameters:
\begin{itemize}

    \item \textbf{Batch Size}: Smaller or larger batch sizes can affect the gradient estimation and model generalization. The smaller ones introduce regularization effects, while the larger size stabilizes the training.
    
    \item \textbf{Learning Rate Scheduling}: One may employ dynamic learning rate schedules, e.g., cosine annealing, step decay, that accelerate convergence and avoid getting stuck in poor local minima.
    
    \item \textbf{Weight Initialization}: Initializing weights with techniques like He or Xavier initialization might improve convergence rates and prevent vanishing/exploding gradients.
    
    \item \textbf{Regularization Techniques}: Adding L2 regularization (weight decay) or using different dropout rates across layers can further mitigate overfitting.

    \item \textbf{Data Augmentation}: Expanding the dataset through techniques like rotation, flipping, and cropping, especially in models like MobileNetV2 or ResNet50, could improve robustness to variability in input data.

    \item \textbf{Epochs and Early Stopping}: Testing different numbers of epochs and introducing early stopping based on validation performance can prevent overfitting while ensuring sufficient learning.

    \item \textbf{Loss Functions}: Exotic loss functions like focal loss may enhance performance when the dataset is highly imbalance-skewed.
    
    \item \textbf{Layer Width and Depth}: The number of neurons per layer or the total number of layers in the CNN might be tuned for feature extraction capabilities.

    \item \textbf{Optimizer Variants}: Trying RMSProp or SGD with momentum may provide performance gains depending on the dataset characteristics.
    
\end{itemize}

Experiments with these parameters would be needed to determine whether they improve accuracy, robustness, or generalization to unseen data.

\subsection{\textbf{Evaluation Metrics}}
To evaluate model performance, we used several key metrics:
\begin{itemize}
    \item \textbf{Accuracy:} The proportion of correct predictions made by the model.
    \item \textbf{Precision:} The proportion of true positive predictions among all positive predictions made by the model.
    \item \textbf{Recall:} The proportion of true positive predictions among all actual positive cases.
    \item \textbf{F1-Score:} The harmonic mean of precision and recall, providing a balance between the two.
    \item \textbf{SSIM (Structural Similarity Index):} A measure of structural similarity between two images; numerical SSIM results are not reported here, and similarity alone does not establish diagnostic realism.
\end{itemize}

\section{\textbf{Results and Discussion}}

The reported classification results compare CNN, MobileNetV2, ResNet50, and VGG19 in terms of accuracy, F1-score, precision, and recall. Table~\ref{tab:classification_reports} reports performance on 300 images, with class 0 representing Normal and class 1 representing Pneumonia. The precision, recall, and F1-score in Table~\ref{tab:performance_metrics} refer to the pneumonia class. These tables describe classifier performance but do not provide a matched real-data versus GAN-augmented comparison for each model.

MobileNetV2 achieved the highest reported accuracy of 88\%, while the CNN achieved the highest pneumonia recall of 92.67\% with precision of 79.43\%. The CNN confusion matrix shows 11 missed pneumonia cases and 36 normal cases classified as pneumonia, compared with 19 and 17, respectively, for MobileNetV2. These results illustrate a sensitivity--precision trade-off. They do not establish that GAN augmentation improved performance or that the models are suitable for real-time clinical use.

\begin{figure}[htbp]
    \centering
    \includegraphics[width=0.5\textwidth]{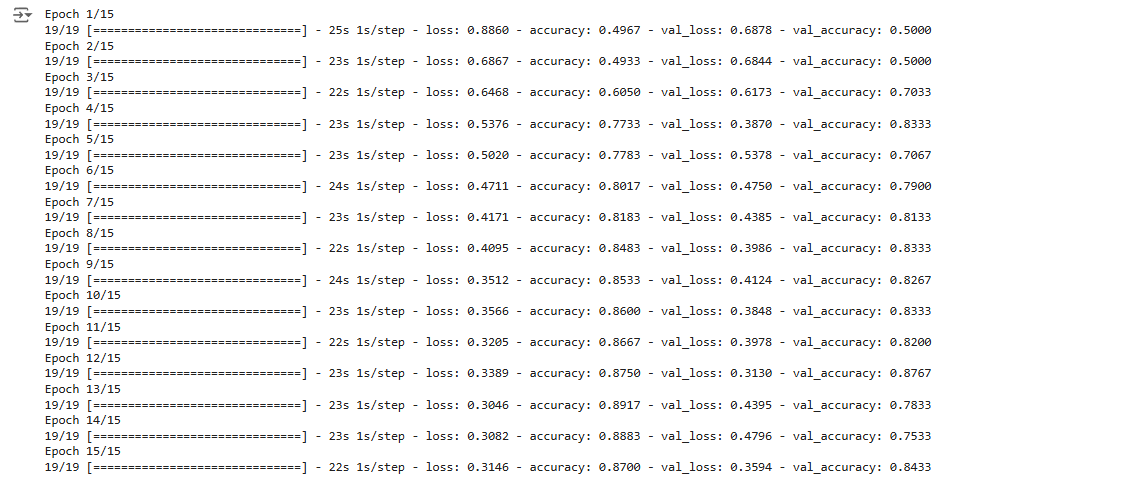}
    \caption{Training Epoch For CNN model}
    \label{fig:cnn-training-epoch}
\end{figure}

\begin{figure}[htbp]
    \centering
    \includegraphics[width=0.5\textwidth]{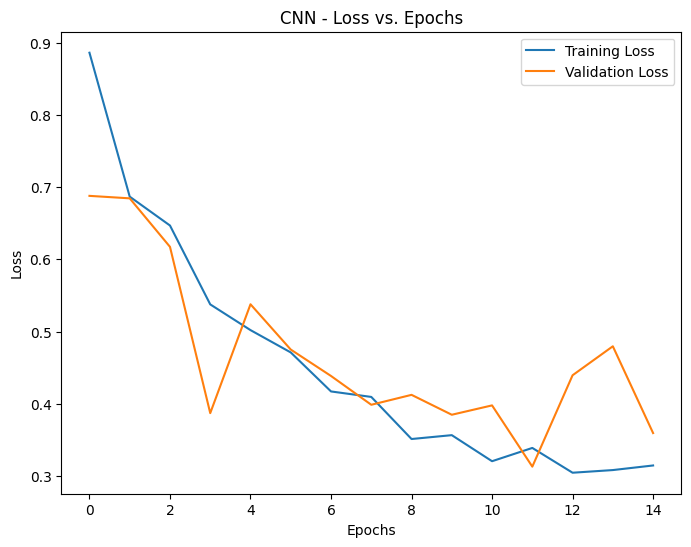}
    \caption{CNN - Loss VS Epoch}
    \includegraphics[width=0.5\textwidth]{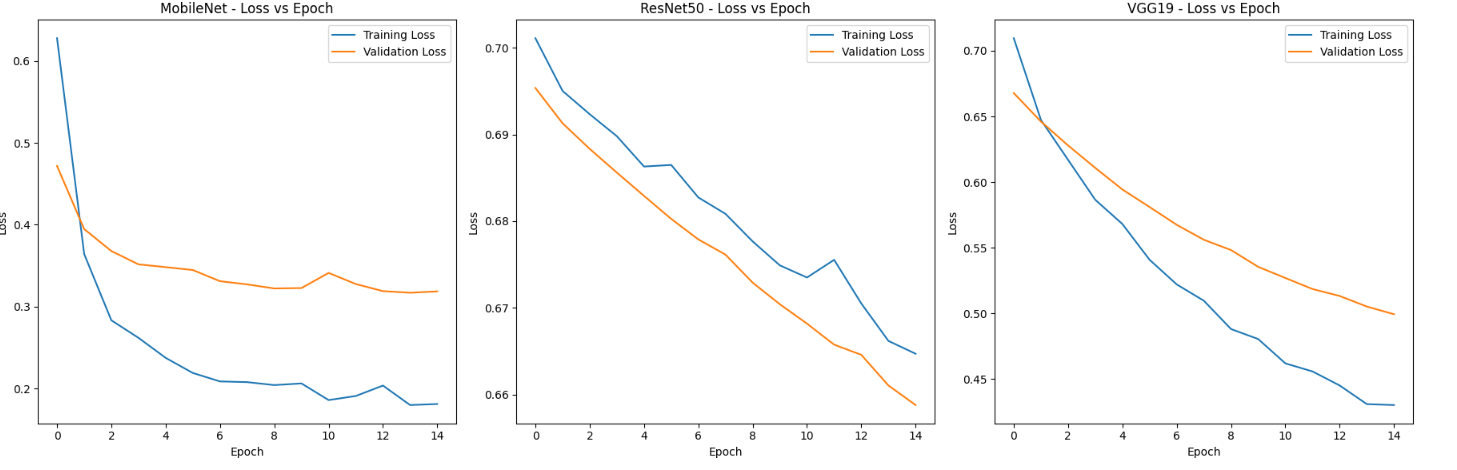}
    \caption{Loss VS Epoch for MobileNet, Resnet50 and VGG19 }
    \label{fig:loss-vs-epoch-all}
\end{figure}

\begin{table}[ht]
\centering
\caption{Classification Reports for Various Models}
\scriptsize
\setlength{\tabcolsep}{5pt} 
\begin{tabular}{lcccccc}
\toprule
\textbf{Model} & \textbf{Class} & \textbf{Precision} & \textbf{Recall} & \textbf{F1-Score} & \textbf{Support} \\
\midrule
\multirow{5}{*}{CNN} & 0 & 0.91 & 0.76 & 0.83 & 150 \\
                     & 1 & 0.79 & 0.93 & 0.86 & 150 \\
                     & \textbf{Acc.} & \multicolumn{4}{c}{0.84 (300)} \\
                     & Macro Avg & 0.85 & 0.84 & 0.84 & 300 \\
                     & Wtd. Avg  & 0.85 & 0.84 & 0.84 & 300 \\
\midrule
\multirow{5}{*}{MobileNetV2} & 0 & 0.88 & 0.89 & 0.88 & 150 \\
                             & 1 & 0.89 & 0.87 & 0.88 & 150 \\
                             & \textbf{Acc.} & \multicolumn{4}{c}{0.88 (300)} \\
                             & Macro Avg & 0.88 & 0.88 & 0.88 & 300 \\
                             & Wtd. Avg  & 0.88 & 0.88 & 0.88 & 300 \\
\midrule
\multirow{5}{*}{ResNet50} & 0 & 0.79 & 0.73 & 0.76 & 150 \\
                          & 1 & 0.75 & 0.81 & 0.78 & 150 \\
                          & \textbf{Acc.} & \multicolumn{4}{c}{0.77 (300)} \\
                          & Macro Avg & 0.77 & 0.77 & 0.77 & 300 \\
                          & Wtd. Avg  & 0.77 & 0.77 & 0.77 & 300 \\
\midrule
\multirow{5}{*}{VGG19} & 0 & 0.84 & 0.79 & 0.82 & 150 \\
                       & 1 & 0.81 & 0.85 & 0.83 & 150 \\
                       & \textbf{Acc.} & \multicolumn{4}{c}{0.82 (300)} \\
                       & Macro Avg & 0.82 & 0.82 & 0.82 & 300 \\
                       & Wtd. Avg  & 0.82 & 0.82 & 0.82 & 300 \\
\bottomrule
\end{tabular}
\label{tab:classification_reports}
\end{table}

 
\begin{table}[ht]
\centering
\caption{Performance Metrics for Various Models}
\begin{tabular}{lcccc}
\toprule
\textbf{Model}       & \textbf{Accuracy} & \textbf{F1 Score} & \textbf{Precision} & \textbf{Recall} \\
\midrule
CNN                  & 0.8433            & 0.8554            & 0.7943             & 0.9267          \\
MobileNet            & 0.8800            & 0.8792            & 0.8851             & 0.8733          \\
ResNet50             & 0.7667            & 0.7756            & 0.7469             & 0.8067          \\
VGG19                & 0.8233            & 0.8285            & 0.8050             & 0.8533          \\
\bottomrule
\end{tabular}
\label{tab:performance_metrics}
\end{table}

\begin{figure}[htbp]
    \centering
    \includegraphics[width=0.5\textwidth]{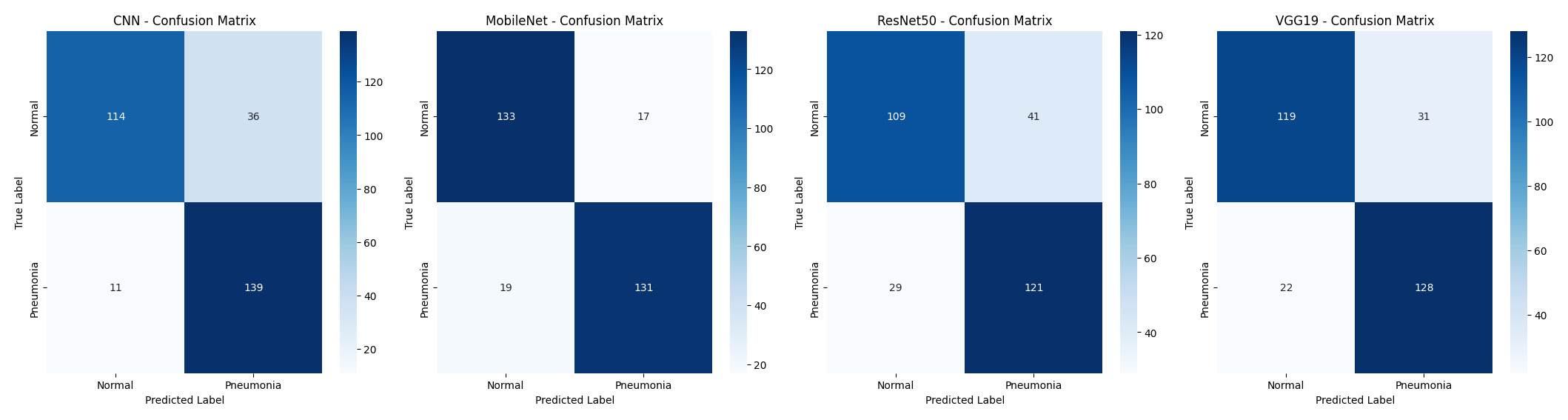}
    \caption{Confusion Matrix for Various Models}
\end{figure}


\subsection{\textbf{Experimental Analysis}}

The training curves illustrate differences between training and validation performance. In the VGG19 real-data versus augmented-data comparison, augmented training accuracy reaches approximately 100\%, while validation accuracy remains near 50\% after the initial epoch. Real-data validation accuracy remains approximately 75--79\%. Thus, the displayed experiment does not show a validation-performance improvement from GAN augmentation. The large training--validation gap is consistent with poor generalization; the available outputs do not establish its cause.

\begin{figure}[htbp]
    \centering
    \includegraphics[width=0.5\textwidth]{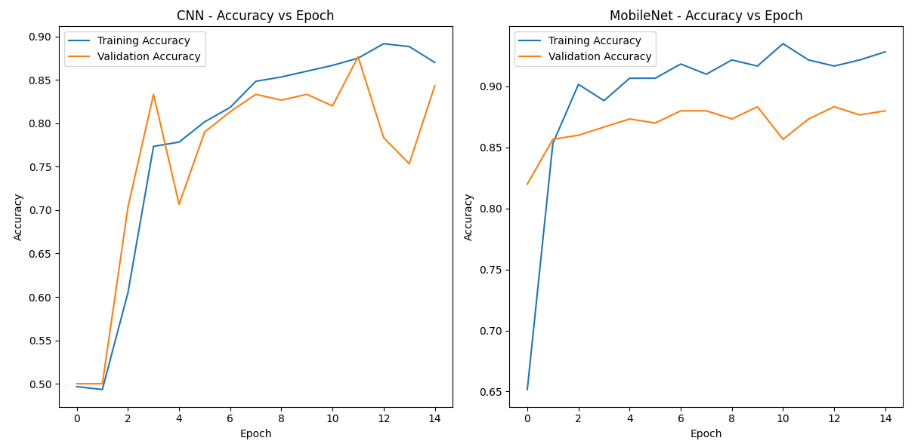}
    \caption{Accuracy VS Epoch for CNN and MobileNet}
    \includegraphics[width=0.5\textwidth]{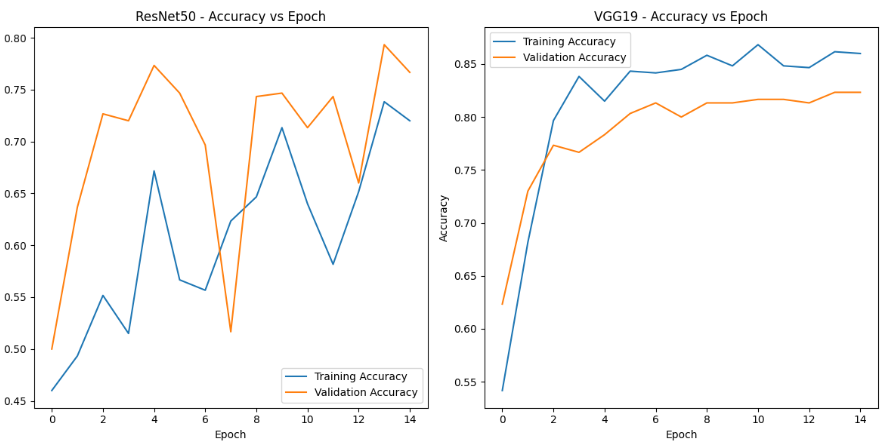}
    \caption{Accuracy versus epoch for ResNet50 and VGG19}
\end{figure}

\begin{figure}[htbp]
    \centering
    \includegraphics[width=0.5\textwidth]{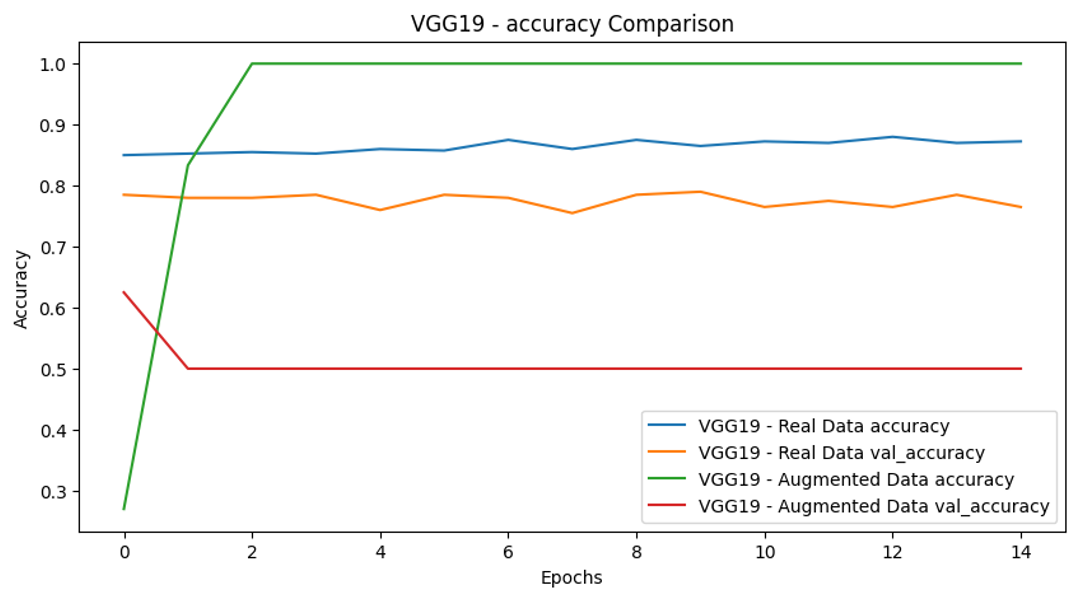}
    \caption{VGG19 training and validation accuracy with real data and GAN-augmented data.}
\end{figure}

\subsubsection{\textbf{Interpretation of ROC Curve and Precision-Recall Curve}}

An ROC curve is a graph illustrating the ability of a model to separate classes at every threshold of possible cut-offs. It plots TPR sensitivity-the True Positive Rate against FPR, the False Positive Rate at various thresholds. AUC-ROC calculates how the model generally tells the differences between positive and negative classes, and values closer to 1 indicate better performance in the model. A steep initial rise in the ROC curve suggests strong predictive capability with minimal false positives at lower thresholds. Conversely, a curve closer to the diagonal line (AUC-ROC $\approx 0.5$) signifies a model with no discriminatory power. The ROC curve is particularly useful when the dataset has a balanced class distribution since it evaluates performance over all thresholds without focusing solely on one metric like precision or recall.

\begin{figure}[htbp]
    \centering
    \includegraphics[width=0.5\textwidth]{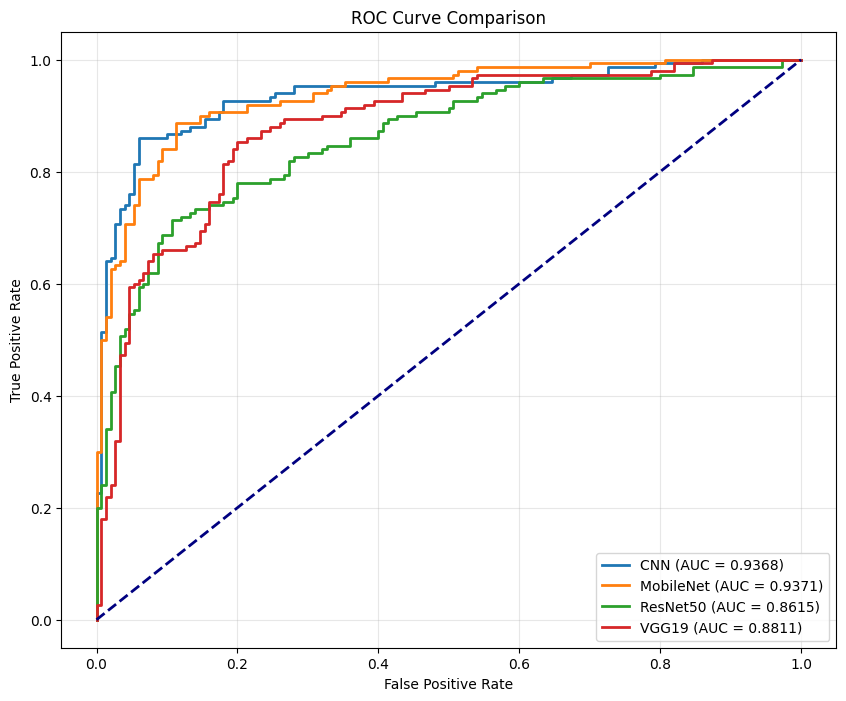}
    \caption{ROC Curve Comparison}
\end{figure}

\begin{figure}[htbp]
    \centering
    \includegraphics[width=0.5\textwidth]{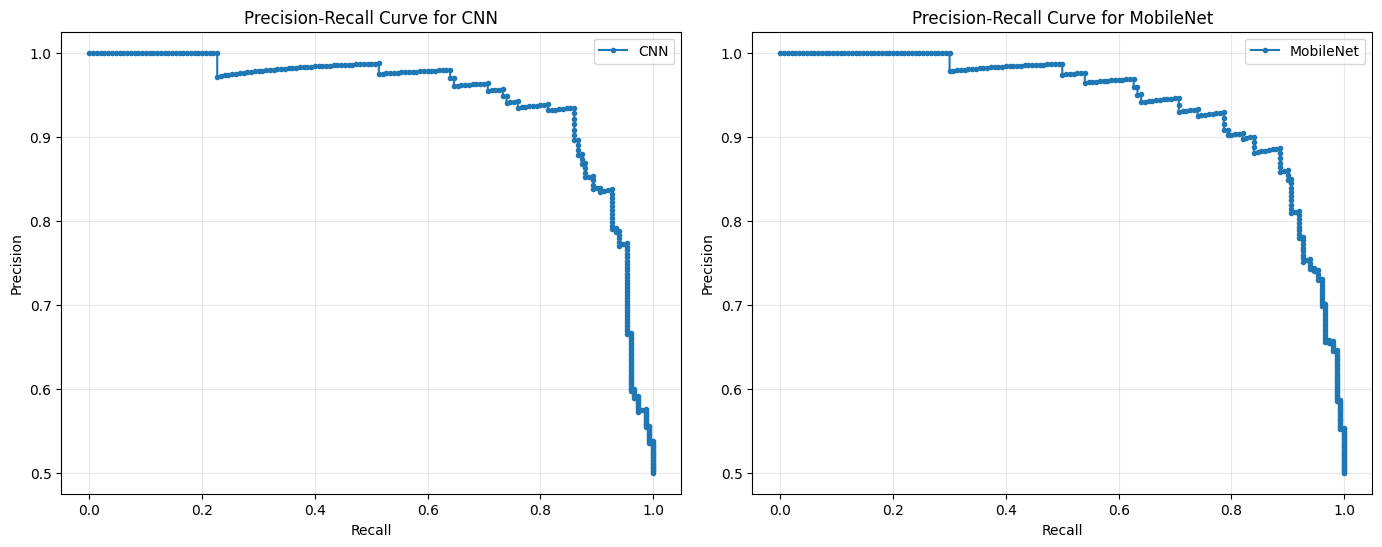}
    \caption{Precision Recall Curve for CNN and MobileNet}
\end{figure}

\begin{figure}[htbp]
    \centering
    \includegraphics[width=0.5\textwidth]{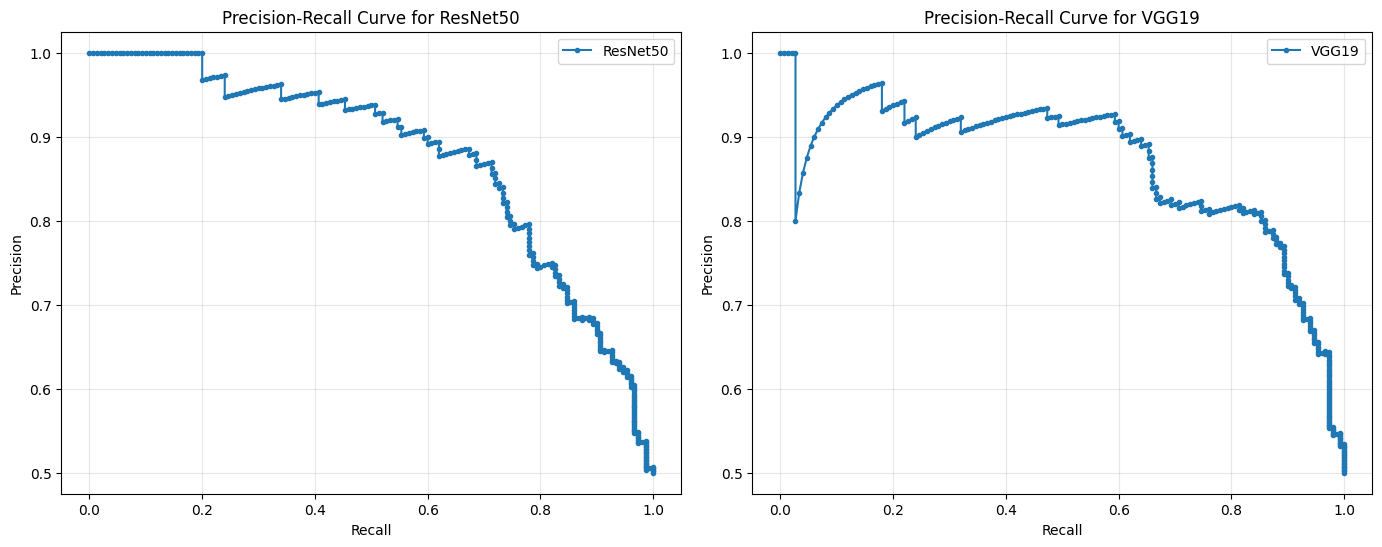}
    \caption{Precision Recall Curve for Resnet50 and VGG19}
\end{figure}

The Precision-Recall Curve plots the trade-off between Precision-Predictive Positive Value and Recall-sensitivity for the positive class. It is very useful in cases of imbalance datasets. Precision tells us what proportion of the predicted positives actually corresponds to true positives, whereas recall tells us what proportion of all actual positives is included. It plots the Precision against Recall at different thresholds, and the area under the PR curve, AUC-PR, is an aggregate measure of performance. A model with high precision and low recall is highly selective in predicting positive outcomes, while a model with high recall but low precision will classify many negatives as positives. A higher PR curve represents better performance, especially in cases where the model effectively balances the two metrics.

1. True Positive Rate (TPR) or Recall:  
   \[
   \text{TPR} = \frac{\text{TP}}{\text{TP} + \text{FN}}
   \]  
   Where TP = True Positives, FN = False Negatives.

2. False Positive Rate (FPR):  
   \[
   \text{FPR} = \frac{\text{FP}}{\text{FP} + \text{TN}}
   \]  
   Where FP = False Positives, TN = True Negatives.

3. Precision:  
   \[
   \text{Precision} = \frac{\text{TP}}{\text{TP} + \text{FP}}
   \]

4. Recall (Sensitivity):  
   \[
   \text{Recall} = \frac{\text{TP}}{\text{TP} + \text{FN}}
   \]

The ROC curve uses TPR and FPR across thresholds, emphasizing the model's balance between sensitivity and specificity. In contrast, the PR curve highlights Precision and Recall, providing insight into a model's positive prediction reliability, particularly for datasets where one class dominates. Understanding these curves together helps determine the most appropriate threshold and model suitability for specific applications.

\section{\textbf{Conclusion}}

This work compares deep learning models for pneumonia classification in chest X-rays and explores GAN-based augmentation. MobileNetV2 achieved the highest reported classification accuracy, while the custom CNN achieved higher pneumonia recall at the cost of more false positives. However, the VGG19 augmentation comparison showed lower validation accuracy with GAN-augmented data than with real data. The presented results therefore do not demonstrate that the implemented GAN augmentation improved generalization.

The synthetic-image example and the VGG19 training--validation gap indicate limitations that require further investigation. Conditional GANs or auxiliary classifier GANs could be explored, but any benefit would need to be demonstrated experimentally. Larger and more diverse datasets, clearer reporting of the evaluation protocol, and external validation are needed before drawing conclusions about clinical applicability.

The custom CNN achieved pneumonia recall of 92.67\%, while MobileNetV2 achieved accuracy of 88\% with balanced class-wise metrics. Accuracy, F1-score, precision, recall, confusion matrices, and training curves describe the observed performance and its limitations. Extending GAN training to 500 epochs remains a proposed experiment, rather than an established route to improved image quality or pneumonia recall. The study supports further investigation of augmentation quality and training settings before claiming a benefit from synthetic data.

\section{Future Work}

Future work will investigate whether refining the GAN can reduce visible artifacts and improve synthetic-image quality. This includes exploring conditional GANs or auxiliary classifier GANs and evaluating whether their outputs improve classification. Learning rates, batch sizes, training duration, and regularization could also be varied for VGG19, MobileNetV2, ResNet50, and the custom CNN. Matched comparisons of real-data and augmented training, together with evaluation on larger and more diverse datasets, are needed to determine whether any changes improve generalization and support clinical applicability.

Beyond binary classification, future evaluation could examine lesion localization and image-grounded question answering. Jha et al. \cite{jha2026dentiaskvqabenchmarkmultimodal} introduced DentiAsk for dental radiographs and found that recognition performance did not translate into comparable localization and counting performance. This distinction motivates evaluating each additional capability separately if the pneumonia system is extended to multimodal interpretation.

If such an extension uses an LLM-powered browser or assistant to process external reports, its evaluation should also address prompt injection. Adhikari et al. \cite{adhikari2026securing} review these threats and defenses such as instruction isolation and provenance tracking, providing relevant considerations for that prospective interface.

\bibliographystyle{IEEEtran}
\bibliography{references}

\end{document}